\documentclass[12pt,a4paper]{article}

\usepackage[utf8]{inputenc}
\usepackage[margin=1in]{geometry}

\usepackage{amsmath}

\usepackage{newtxtext,newtxmath} 

\usepackage{graphicx}
\usepackage{booktabs}
\usepackage{multirow}
\usepackage{array}
\usepackage{caption}
\usepackage{subcaption}
\usepackage{float}
\usepackage{xcolor}

\usepackage{natbib}
\usepackage{hyperref}

\usepackage{setspace}
\usepackage[protrusion=true,expansion=true,verbose=silent]{microtype}
\microtypesetup{nopatch=footnote}
\hypersetup{
    colorlinks=true,
    linkcolor=blue,
    citecolor=blue,
    urlcolor=blue,
}

\title{\textbf{Pediatric Pneumonia Detection from Chest X-Rays: A Comparative Study of Transfer Learning and Custom CNNs}}

\author{
    Agniv Roy Choudhury\\
    Department of Computer Science\\
    University of Texas at Austin
}

\date{November 2025}

\begin{document}

\maketitle

\begin{abstract}
\textbf{Background}: Pneumonia remains a leading cause of mortality in children under five years, responsible for over 700{,}000 deaths annually worldwide. Accurate and timely diagnosis from chest X-rays is critical but limited by radiologist availability and inter-observer variability, especially in resource-constrained settings.

\textbf{Objective}: This study compares custom Convolutional Neural Networks (CNNs) trained from scratch with transfer learning approaches using ImageNet-pretrained architectures (ResNet50, DenseNet121, EfficientNet-B0) for automated pediatric pneumonia detection from chest X-rays. Two transfer learning regimes are evaluated - feature extraction (frozen backbone) and fine-tuning (differential learning rates) to identify effective training strategies for medical imaging with limited data.

\textbf{Methods}: A dataset of 5{,}216 pediatric chest X-rays (ages 1 - 5 years) from Guangzhou Women and Children's Medical Center was used, with an 80/10/10 stratified split (4{,}172 train, 521 validation, 523 test) constructed to address the original dataset's inadequate 16-image validation set. Seven models were trained and evaluated: one custom CNN baseline and six transfer learning models (three architectures $\times$ two regimes). Performance was assessed using accuracy, precision, recall, F1-score, AUC, sensitivity, specificity, and confusion matrices. Gradient-weighted Class Activation Mapping (Grad-CAM) was used for explainability.

\textbf{Results}: Fine-tuned ResNet50 achieved the best performance, with 99.43\% accuracy (520/523 correct), 99.61\% F1-score, and 99.93\% AUC. It made only 3 errors (1 false positive, 2 false negatives), improving accuracy by 3.06 percentage points over the custom CNN baseline (96.37\%). Across all architectures, fine-tuning outperformed frozen-backbone training by an average of 5.48 percentage points in accuracy. The best model achieved 99.48\% sensitivity (386/388 pneumonia cases detected) and 99.26\% specificity (134/135 normal cases). Grad-CAM visualizations confirmed that predictions were driven by clinically relevant lung regions and pathological features.

\textbf{Conclusions}: Transfer learning with fine-tuning substantially outperforms CNNs trained from scratch for pediatric pneumonia detection, achieving near-perfect performance with very few errors. The marked performance gap between frozen and fine-tuned models underscores the importance of domain adaptation via differential learning rates in medical imaging. With only 2 missed pneumonia cases out of 388, the proposed system shows strong potential as a screening tool to assist radiologists, particularly in resource-limited settings. Future work should validate these findings on adult populations, multi-center datasets, and more diverse clinical scenarios.

\end{abstract}

\newpage

\section{Introduction}

\subsection{Pneumonia Burden Worldwide}

Pneumonia remains one of the leading causes of morbidity and mortality globally, particularly among children under five years of age and elderly populations. According to the World Health Organization (WHO), pneumonia accounts for approximately 15\% of all deaths in children under five, claiming the lives of over 700,000 children annually \citep{who2022}. In the United States alone, pneumonia results in over 1.5 million emergency department visits and 50,000 deaths each year \citep{cdc2021}. The disease burden is particularly severe in low and middle income countries, where access to timely diagnosis and treatment remains limited \citep{rudan2008}.

Early and accurate diagnosis of pneumonia is critical for effective treatment and improved patient outcomes. Chest X-ray (CXR) imaging is the primary diagnostic tool for pneumonia detection, offering a non-invasive and relatively inexpensive method for visualizing lung abnormalities \citep{franquet2001}.

\subsection{Machine Learning in Medical Imaging}

The advent of deep learning has revolutionized medical image analysis, demonstrating remarkable success in various diagnostic tasks including disease detection, classification, and segmentation \citep{litjens2017survey}. Convolutional Neural Networks (CNNs), in particular, have shown human-level or superior performance in analyzing medical images, including chest X-rays, CT scans, and MRI images \citep{esteva2017}. However, the interpretation of chest X-rays requires significant expertise and can be subject to inter-observer variability, with reported agreement rates between radiologists ranging from 60\% to 80\% \citep{neuman2010}. This variability, combined with the shortage of trained radiologists in many regions, creates a pressing need for automated diagnostic tools \citep{mollura2014}.

Transfer learning, a technique that leverages knowledge learned from large-scale datasets (such as ImageNet) and adapts it to specific medical imaging tasks, has emerged as a particularly promising approach \citep{tajbakhsh2016}. Pre-trained models like ResNet, DenseNet, and EfficientNet have shown superior performance compared to models trained from scratch, especially when medical imaging datasets are limited in size \citep{shin2016}. The ability to fine-tune these models with domain-specific data allows them to capture both general visual features and task-specific patterns, potentially leading to more robust and accurate diagnostic systems \citep{raghu2019}.

\subsection{Research Gap}

Despite the promising results of deep learning in pneumonia detection, several research gaps remain:

\begin{enumerate}
    \item \textbf{Limited Comparative Studies}: While numerous studies have explored either custom CNN architectures or transfer learning approaches independently, comprehensive comparisons between these methodologies using identical datasets and evaluation protocols are scarce \citep{chouhan2020}.
    
    \item \textbf{Model Explainability}: Many existing studies focus solely on performance metrics without providing insights into model decision-making, which is crucial for clinical adoption and trust \citep{holzinger2017}.
    
    \item \textbf{Pediatric Population Focus}: Most pneumonia detection studies focus on adult populations, with limited research on pediatric chest X-rays, which present unique challenges due to anatomical differences and image characteristics \citep{jain2020}.
\end{enumerate}

\subsection{Research Questions and Hypotheses}

This study addresses the following research questions:

\textbf{RQ1}: How does transfer learning compare to training CNNs from scratch for pediatric pneumonia detection from chest X-rays?

\textbf{RQ2}: What is the impact of different training regimes (feature extraction vs. fine-tuning) on transfer learning model performance?

\textbf{RQ3}: Which deep learning architecture (ResNet50, DenseNet121, or EfficientNet-B0) achieves the best performance for pneumonia detection?

\textbf{RQ4}: Where do models focus their attention when making predictions, and how does this relate to clinical interpretability?

\textbf{Hypotheses}:

\textbf{H1}: Transfer learning models will achieve higher accuracy and F1-scores compared to custom CNNs trained from scratch, due to leveraging pre-trained ImageNet features.

\textbf{H2}: Fine-tuning strategies (unfreezing deeper layers with differential learning rates) will outperform feature extraction approaches (frozen backbone) by allowing domain-specific adaptation.

\textbf{H3}: Modern architectures (DenseNet121, EfficientNet-B0) will demonstrate competitive or superior performance compared to ResNet50 while requiring fewer parameters.

\textbf{H4}: Grad-CAM visualizations will reveal that high-performing models focus on clinically relevant regions (lung fields, infiltrates) rather than spurious correlations.

\subsection{Contributions}

This research makes the following contributions:

\begin{enumerate}
    \item \textbf{Comprehensive Comparison}: Provided a systematic comparison of custom CNN architectures versus three state-of-the-art transfer learning models (ResNet50, DenseNet121, EfficientNet-B0) on pediatric pneumonia detection.
    
    \item \textbf{Training Regime Analysis}: Evaluated the impact of different transfer learning strategies (feature extraction vs. fine-tuning with differential learning rates) on model performance.
    
    \item \textbf{Rigorous Methodology}: Addressed the original dataset's inadequate validation set (16 images) by creating a proper 80/10/10 stratified split, ensuring reliable model selection and evaluation.
    
    \item \textbf{Model Explainability}: Implemented Grad-CAM (Gradient-weighted Class Activation Mapping) to visualize model attention and provide clinical interpretability for all prediction categories (true positives, true negatives, false positives, false negatives).
    
    \item \textbf{Clinical Metrics}: Reported comprehensive clinical metrics including sensitivity, specificity, positive predictive value (PPV), and negative predictive value (NPV), in addition to standard machine learning metrics.
    
    \item \textbf{Reproducible Research}: Provided detailed documentation of our methodology, including data split rationale, hyperparameters, and training procedures, facilitating reproducibility and future research.
\end{enumerate}

\subsection{Paper Organization}

The remainder of this paper is organized as follows: Section 2 reviews related work in deep learning for medical imaging and pneumonia detection. Section 3 describes methodologies, including dataset preparation, model architectures, training procedures, and evaluation metrics. Section 4 presents our experimental results with detailed performance comparisons. Section 5 discusses the findings, clinical implications, and limitations. Section 6 concludes the paper and outlines future research directions.
\section{Related Work}

\subsection{Deep Learning in Medical Imaging}

Deep learning has transformed medical image analysis over the past decade, with convolutional neural networks (CNNs) demonstrating remarkable capabilities in classification, detection, and segmentation across multiple imaging modalities \citep{litjens2017survey}. The ImageNet Large Scale Visual Recognition Challenge (ILSVRC) \citep{russakovsky2015} catalyzed the development of increasingly sophisticated CNN architectures, many of which have been successfully adapted for medical imaging tasks.

\subsection{CNN Architectures for Image Classification}

Several landmark CNN architectures have shaped the field:

\textbf{ResNet} (Residual Networks): He et al. \citep{he2016} introduced residual connections that enable training of very deep networks by addressing the vanishing gradient problem. ResNet50 has become a standard baseline for transfer learning due to its balance between depth and computational efficiency.

\textbf{DenseNet} (Densely Connected Networks): Huang et al. \citep{huang2017} proposed dense connections where each layer receives input from all preceding layers, promoting feature reuse and reducing parameters. DenseNet121 has shown particular promise due to its parameter efficiency.

\textbf{EfficientNet}: Tan and Le \citep{tan2019} introduced a compound scaling method that uniformly scales network depth, width, and resolution. EfficientNet-B0 offers an attractive trade-off between performance and computational cost.

\subsection{Transfer Learning in Medical Imaging}

Transfer learning has emerged as a dominant paradigm in medical image analysis, particularly when labeled data is limited. Tajbakhsh et al. \citep{tajbakhsh2016} demonstrated that ImageNet pre-trained CNNs often outperform models trained from scratch, while Raghu et al. \citep{raghu2019} found that transfer learning benefits depend on the target task and dataset size. Two primary strategies exist: \textbf{feature extraction} (freezing convolutional layers, training only the classifier) \citep{donahue2014} and \textbf{fine-tuning} (unfreezing layers with lower learning rates for domain adaptation) \citep{yosinski2014}.

\subsection{Pneumonia Detection Using Deep Learning}

Several studies have applied deep learning to pneumonia detection from chest X-rays:

\textbf{CheXNet}: Rajpurkar et al. \citep{rajpurkar2017} developed a 121-layer DenseNet model that achieved radiologist-level performance on pneumonia detection, demonstrating 0.7632 AUC on the ChestX-ray14 dataset. Their work highlighted the potential of deep learning to match expert-level diagnosis.

\textbf{Pediatric Pneumonia Detection}: Kermany et al. \citep{kermany2018} created a large dataset of pediatric chest X-rays and trained a custom CNN achieving 92.8\% accuracy in distinguishing normal from pneumonia cases. This dataset has become a benchmark for pediatric pneumonia detection research.

\textbf{Ensemble Approaches}: Stephen et al. \citep{stephen2019} explored ensemble methods combining multiple CNN architectures, achieving 95.3\% accuracy on pneumonia detection. However, ensemble approaches increase computational complexity and deployment challenges.

\textbf{Attention Mechanisms}: Guan et al. \citep{guan2019} incorporated attention mechanisms into CNN architectures for pneumonia detection, improving both performance and interpretability by highlighting relevant image regions.

\subsection{Model Explainability in Medical AI}

The ``black box'' nature of deep learning models has raised concerns in clinical applications, where interpretability is crucial for trust and adoption \citep{caruana2015}. \textbf{Grad-CAM} (Gradient-weighted Class Activation Mapping) \citep{selvaraju2017} uses gradients flowing into the final convolutional layer to produce localization maps highlighting important regions, and has been widely adopted in medical imaging for providing visual explanations without modifying model architecture. Other techniques include saliency maps \citep{simonyan2013} and Layer-wise Relevance Propagation (LRP) \citep{bach2015}, though these can be noisy or computationally intensive.

\subsection{Challenges in Medical Imaging Datasets}

Medical imaging datasets present unique challenges including class imbalance \citep{johnson2019}, limited dataset size due to privacy concerns and annotation costs \citep{willemink2020}, domain shift across institutions \citep{zech2018}, and inadequate validation sets that hinder reliable model selection \citep{varoquaux2022}.

\subsection{Research Gaps Addressed}

While existing research has made significant progress in pneumonia detection, this work addresses several gaps:

\begin{enumerate}
    \item \textbf{Systematic Comparison}: Most studies focus on a single architecture or approach, lacking comprehensive comparisons across multiple state-of-the-art models under identical conditions.
    
    \item \textbf{Training Regime Analysis}: Limited research has systematically compared feature extraction versus fine-tuning strategies with differential learning rates for pneumonia detection.
    
    \item \textbf{Validation Set Adequacy}: Addressed the original dataset's inadequate validation set (16 images) by creating a proper stratified split, ensuring reliable model evaluation.
    
    \item \textbf{Comprehensive Explainability}: Provided Grad-CAM visualizations for all prediction categories (TP, TN, FP, FN), offering insights into both correct and incorrect predictions.
    
    \item \textbf{Clinical Metrics}: Reported comprehensive clinical metrics (sensitivity, specificity, PPV, NPV) alongside standard ML metrics, providing a complete picture of clinical utility.
\end{enumerate}
\section{Materials and Methods}

\subsection{Dataset}

\subsubsection{Data Source and Description}

Utilized the Chest X-Ray Images (Pneumonia) dataset from Kaggle, originally collected by Kermany et al. \citep{kermany2018} from Guangzhou Women and Children's Medical Center, Guangzhou, China. The dataset comprises chest X-ray images from pediatric patients aged 1 to 5 years, labeled as either NORMAL or PNEUMONIA by expert physicians.

The original dataset contained 5,856 images distributed across three splits: 5,216 training images, 16 validation images, and 624 test images. However, the validation set of only 16 images (8 normal, 8 pneumonia) was statistically insufficient for reliable model validation, hyperparameter tuning, and early stopping decisions.

\subsubsection{Data Split Rationale}

To address this critical limitation, created a new stratified 80/10/10 split from the original 5,216 training images, resulting in:
\begin{itemize}
    \item \textbf{Training set}: 4,172 images (80\%)
    \item \textbf{Validation set}: 521 images (10\%)
    \item \textbf{Test set}: 523 images (10\%)
\end{itemize}

The split was performed using scikit-learn's \texttt{train\_test\_split} with stratified sampling (random\_state=42) to maintain consistent class distribution across all splits.

\subsubsection{Class Distribution}

\begin{figure}[H]
\centering
\includegraphics[width=0.85\textwidth]{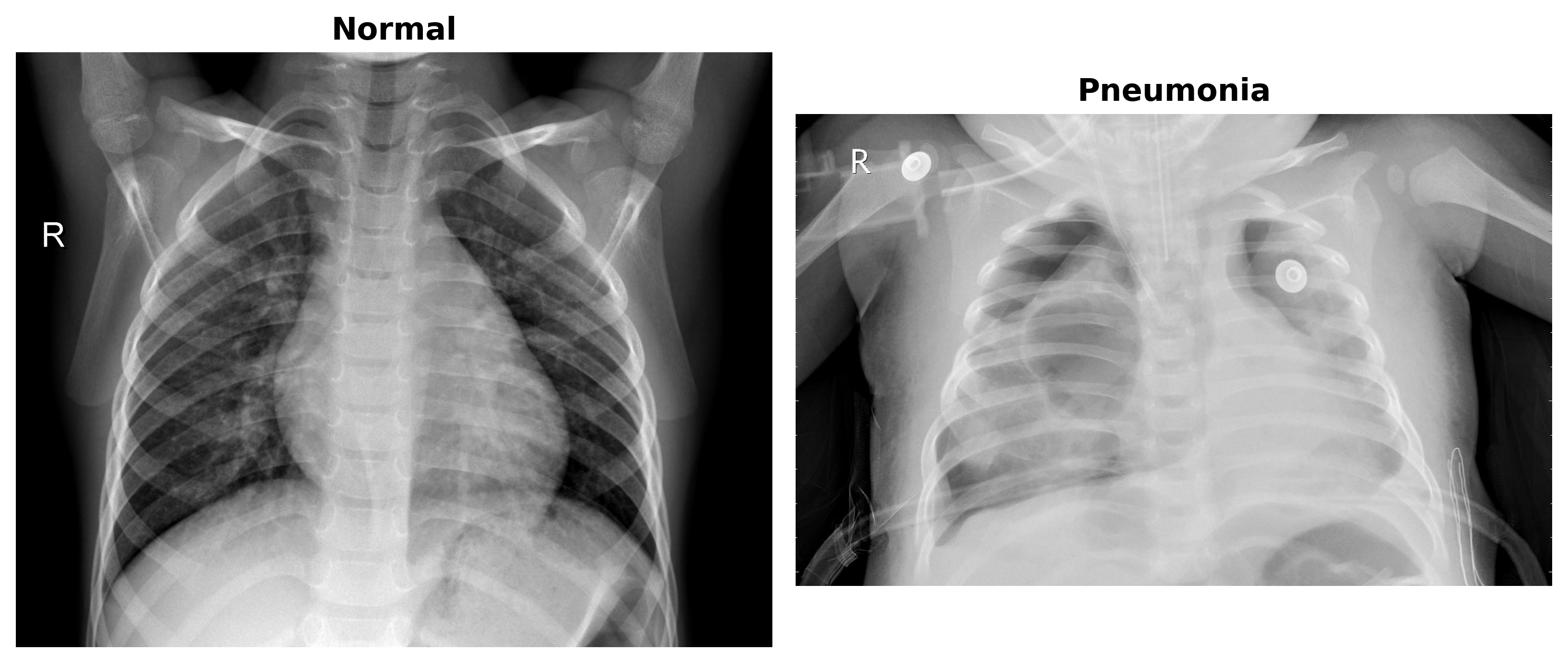}
\caption{Example chest X-ray images from the dataset. Left: Normal case showing clear lung fields. Right: Pneumonia case with visible lung opacities indicating infection.}
\label{fig:sample_xrays}
\end{figure}

The dataset exhibits class imbalance with approximately 74\% pneumonia cases and 26\% normal cases. Table \ref{tab:class_distribution} presents the class distribution across all splits. Figure~\ref{fig:sample_xrays} illustrates representative normal and pneumonia chest X-ray images from the dataset.

\begin{table}[H]
\centering
\caption{Class Distribution Across Dataset Splits}
\label{tab:class_distribution}
\begin{tabular}{lrrrr}
\toprule
\textbf{Split} & \textbf{Normal} & \textbf{Pneumonia} & \textbf{Total} & \textbf{Ratio} \\
\midrule
Train & 1,072 (26\%) & 3,100 (74\%) & 4,172 & 2.89:1 \\
Validation & 134 (26\%) & 387 (74\%) & 521 & 2.89:1 \\
Test & 135 (26\%) & 388 (74\%) & 523 & 2.87:1 \\
\midrule
\textbf{Total} & \textbf{1,341} & \textbf{3,875} & \textbf{5,216} & \textbf{2.89:1} \\
\bottomrule
\end{tabular}
\end{table}

\subsection{Ethical Considerations}

The dataset used in this study is publicly available on Kaggle and is fully de-identified, containing no personally identifiable information. The original dataset creators \citep{kermany2018} obtained institutional approval from Guangzhou Women and Children's Medical Center, and no direct human subject contact was involved in this study. Therefore, this work is exempt from additional IRB review. All data handling and analysis procedures comply with ethical standards for secondary use of de-identified medical imaging data.

\subsection{Data Preprocessing and Augmentation}

\subsubsection{Preprocessing Pipeline}

All images underwent standardized preprocessing:
\begin{enumerate}
    \item \textbf{Resizing}: Images resized to 224$\times$224 pixels
    \item \textbf{Normalization}: Pixel values normalized using ImageNet statistics (mean=[0.485, 0.456, 0.406], std=[0.229, 0.224, 0.225])
    \item \textbf{Format conversion}: Grayscale images converted to RGB format (3 channels)
\end{enumerate}

\subsubsection{Data Augmentation}

Data augmentation was applied exclusively to the training set:
\begin{itemize}
    \item Horizontal flips (50\% probability)
    \item Random rotations ($\pm$10 degrees)
    \item Random affine transformations (translation $\pm$10\%, scale 0.9-1.1$\times$)
    \item Color jitter (brightness and contrast $\pm$20\%)
\end{itemize}

\subsection{Model Architectures}

\subsubsection{Baseline: Custom CNN}

The custom CNN baseline consists of:
\begin{itemize}
    \item 4 convolutional blocks (32$\rightarrow$64$\rightarrow$128$\rightarrow$256 filters)
    \item ReLU activation and MaxPooling (2$\times$2) after each block
    \item Fully connected layers: 50,176 $\rightarrow$ 512 $\rightarrow$ 2 neurons
    \item Dropout (0.5) before final layer
    \item Total parameters: 26 million (100\% trainable)
\end{itemize}

\subsubsection{Transfer Learning Architectures}

Evaluated three ImageNet pre-trained architectures:

\textbf{ResNet50} \citep{he2016}: 50-layer residual network with 23 million parameters.

\textbf{DenseNet121} \citep{huang2017}: 121-layer densely connected network with 7 million parameters.

\textbf{EfficientNet-B0} \citep{tan2019}: Compound-scaled network with 4 million parameters.

\subsubsection{Transfer Learning Strategies}

Two strategies were evaluated for each architecture:

\textbf{Feature Extraction (Frozen Backbone)}:
\begin{itemize}
    \item All pre-trained layers frozen
    \item Only final classification layer trained
    \item Learning rate: 0.001
    \item Trainable parameters: 1-2 million (9-25\%)
\end{itemize}

\textbf{Fine-tuning (Differential Learning Rates)}:
\begin{itemize}
    \item Last two convolutional blocks unfrozen
    \item Backbone LR: 0.0001, Classifier LR: 0.001
    \item Trainable parameters: 2-11 million (43-50\%)
\end{itemize}

\subsection{Training Configuration}

\begin{itemize}
    \item \textbf{Loss function}: CrossEntropyLoss
    \item \textbf{Optimizer}: Adam
    \item \textbf{Batch size}: 32
    \item \textbf{Maximum epochs}: 50
    \item \textbf{Learning rate scheduler}: ReduceLROnPlateau (patience=5, factor=0.5)
    \item \textbf{Early stopping}: Patience=10 epochs
    \item \textbf{Hardware}: Google Colab with NVIDIA A100 GPU
    \item \textbf{Framework}: PyTorch 2.0
\end{itemize}

\subsection{Evaluation Metrics}

\subsubsection{Classification Metrics}
\begin{itemize}
    \item Accuracy, Precision, Recall, F1-Score
    \item Area Under ROC Curve (AUC)
\end{itemize}

\subsubsection{Clinical Metrics}
\begin{itemize}
    \item Sensitivity (True Positive Rate)
    \item Specificity (True Negative Rate)
    \item Positive Predictive Value (PPV)
    \item Negative Predictive Value (NPV)
\end{itemize}

\subsubsection{Confusion Matrix Analysis}
\begin{itemize}
    \item True Positives (TP): Correctly identified pneumonia
    \item True Negatives (TN): Correctly identified normal
    \item False Positives (FP): Normal misclassified as pneumonia
    \item False Negatives (FN): Pneumonia misclassified as normal
\end{itemize}

\subsection{Model Explainability}

Implemented Gradient-weighted Class Activation Mapping (Grad-CAM) \citep{selvaraju2017} to visualize model attention. Grad-CAM generates heatmaps by:
\begin{enumerate}
    \item Computing gradients of predicted class w.r.t. final convolutional layer
    \item Global average pooling of gradients to obtain importance weights
    \item Weighted combination of activation maps
    \item ReLU activation and normalization
\end{enumerate}

Visualizations were generated for four categories: True Positives, True Negatives, False Positives, and False Negatives (4 examples each per model).

\subsection{Ensemble Methods}

Three ensemble strategies were evaluated:
\begin{itemize}
    \item \textbf{Simple Averaging}: Equal-weight average of prediction probabilities
    \item \textbf{Weighted Averaging}: F1-score weighted average
    \item \textbf{Majority Voting}: Majority vote of predicted classes
\end{itemize}

\subsection{Statistical Analysis}

All experiments used fixed random seed (42) for reproducibility. Performed:
\begin{itemize}
    \item Model performance comparison across architectures
    \item Frozen vs fine-tuned analysis
    \item Baseline vs transfer learning comparison
    \item Sensitivity vs specificity trade-off analysis
    \item Class-wise performance evaluation
    \item Failure case analysis
\end{itemize}
\section{Results}

\subsection{Overall Model Performance}

Trained and evaluated seven models: one custom CNN baseline and six transfer learning models. Table \ref{tab:overall_performance} presents the comprehensive performance comparison on the test set (n=523).

\begin{table}[H]
\centering
\caption{Overall Model Performance Comparison}
\label{tab:overall_performance}
\small
\begin{tabular}{llrrrrrrr}
\toprule
\textbf{Model} & \textbf{Mode} & \textbf{Acc} & \textbf{Prec} & \textbf{Rec} & \textbf{F1} & \textbf{AUC} & \textbf{Sens} & \textbf{Spec} \\
\midrule
\textbf{ResNet50} & \textbf{Finetune} & \textbf{99.43} & \textbf{99.74} & \textbf{99.48} & \textbf{99.61} & \textbf{99.93} & \textbf{99.48} & \textbf{99.26} \\
DenseNet121 & Finetune & 98.85 & 99.23 & 99.23 & 99.23 & 99.89 & 99.23 & 97.78 \\
Custom CNN & Scratch & 96.37 & 98.17 & 96.91 & 97.54 & 99.23 & 96.91 & 94.81 \\
EfficientNet & Finetune & 96.37 & 98.17 & 96.91 & 97.54 & 99.49 & 96.91 & 94.81 \\
DenseNet121 & Frozen & 94.46 & 96.62 & 95.88 & 96.25 & 98.47 & 95.88 & 90.37 \\
ResNet50 & Frozen & 92.93 & 95.12 & 95.36 & 95.24 & 97.71 & 95.36 & 85.93 \\
EfficientNet & Frozen & 90.82 & 98.57 & 88.92 & 93.50 & 98.28 & 88.92 & 96.30 \\
\bottomrule
\end{tabular}
\end{table}

ResNet50 with fine-tuning achieved the highest performance across all metrics, with 99.43\% accuracy and 99.61\% F1-score. Notably, this model made only 3 errors out of 523 test images: 1 false positive (0.19\%) and 2 false negatives (0.38\%).

\begin{figure}[H]
\centering
\includegraphics[width=0.95\textwidth]{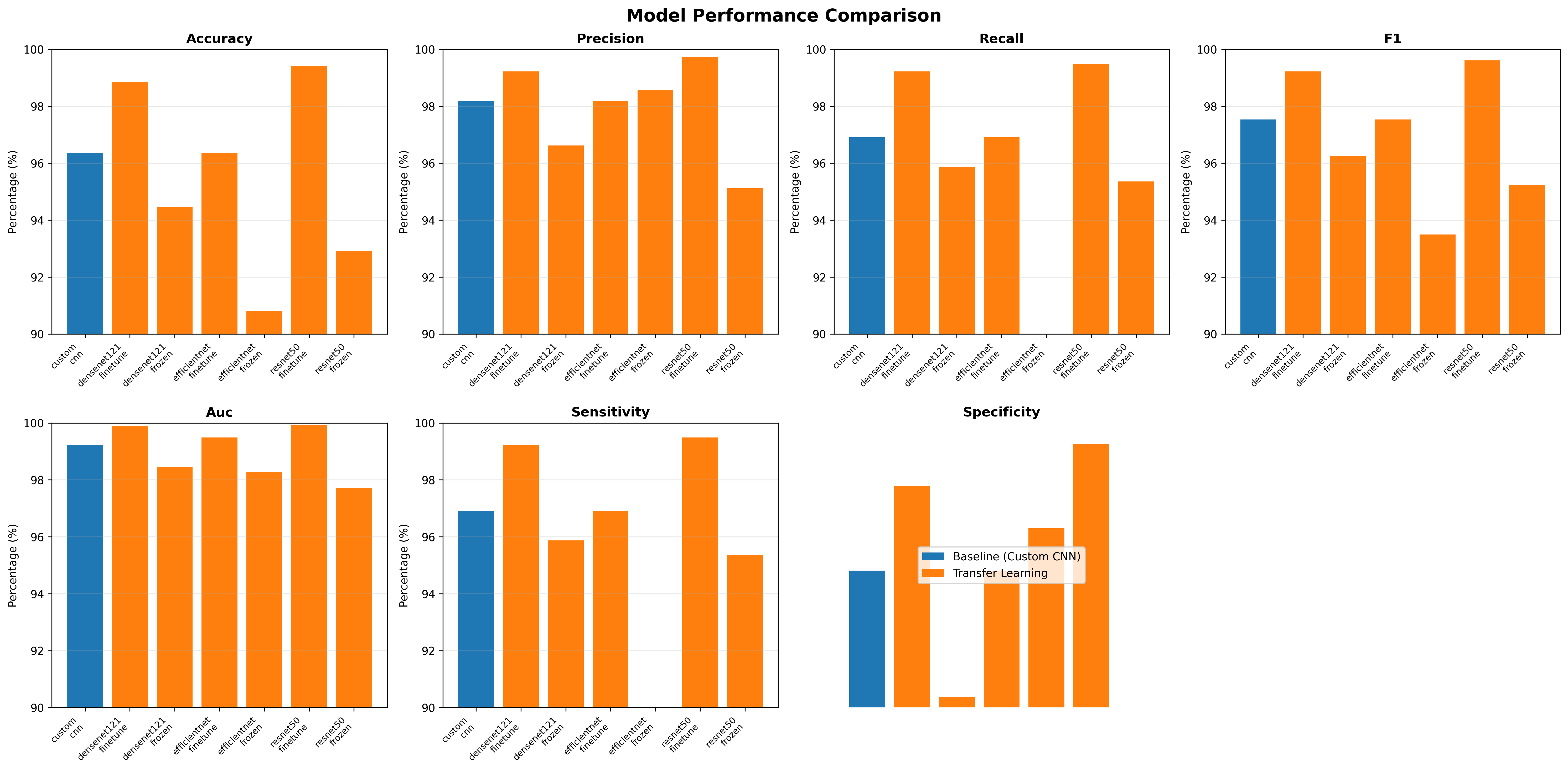}
\caption{Performance metrics comparison across all seven models. Transfer learning with fine-tuning (orange bars) consistently outperforms the baseline (blue bar) and frozen models across all metrics.}
\label{fig:metrics_comparison}
\end{figure}

Figure \ref{fig:metrics_comparison} presents a visual comparison of all model performances across key metrics.

\subsection{Transfer Learning vs Baseline Comparison}

Transfer learning with fine-tuning significantly outperformed the custom CNN baseline. Table \ref{tab:tl_vs_baseline} presents the detailed comparison.

\begin{table}[H]
\centering
\caption{Transfer Learning Improvement Over Baseline}
\label{tab:tl_vs_baseline}
\begin{tabular}{lrrr}
\toprule
\textbf{Metric} & \textbf{Custom CNN} & \textbf{ResNet50 (TL)} & \textbf{Improvement} \\
\midrule
Accuracy (\%) & 96.37 & 99.43 & +3.06 \\
F1-Score (\%) & 97.54 & 99.61 & +2.08 \\
AUC (\%) & 99.23 & 99.93 & +0.70 \\
Sensitivity (\%) & 96.91 & 99.48 & +2.58 \\
Specificity (\%) & 94.81 & 99.26 & +4.44 \\
Total Errors & 19 & 3 & -84.21\% \\
False Negatives & 12 & 2 & -83.33\% \\
\bottomrule
\end{tabular}
\end{table}

The most substantial improvements were observed in specificity (+4.44\%) and error reduction (84\% fewer errors). Critically, false negatives decreased from 12 to 2, an 83\% reduction.

\subsection{Training Regime Analysis}

Fine-tuning consistently outperformed feature extraction across all architectures. Table \ref{tab:frozen_vs_finetune} presents the comparison.

\begin{table}[H]
\centering
\caption{Fine-tuning vs Frozen Backbone Performance}
\label{tab:frozen_vs_finetune}
\begin{tabular}{lrrrr}
\toprule
\textbf{Architecture} & \textbf{Frozen Acc} & \textbf{Finetune Acc} & \textbf{$\Delta$ Acc} & \textbf{$\Delta$ F1} \\
\midrule
ResNet50 & 92.93 & 99.43 & +6.50 & +4.37 \\
DenseNet121 & 94.46 & 98.85 & +4.39 & +2.98 \\
EfficientNet-B0 & 90.82 & 96.37 & +5.55 & +4.04 \\
\midrule
\textbf{Average} & \textbf{92.74} & \textbf{98.22} & \textbf{+5.48} & \textbf{+3.80} \\
\bottomrule
\end{tabular}
\end{table}

The average improvement from fine-tuning was 5.48\% in accuracy and 3.80\% in F1-score. ResNet50 showed the largest improvement (+6.50\%), demonstrating that deeper architectures benefit more from fine-tuning.

\begin{figure}[H]
\centering
\includegraphics[width=0.85\textwidth]{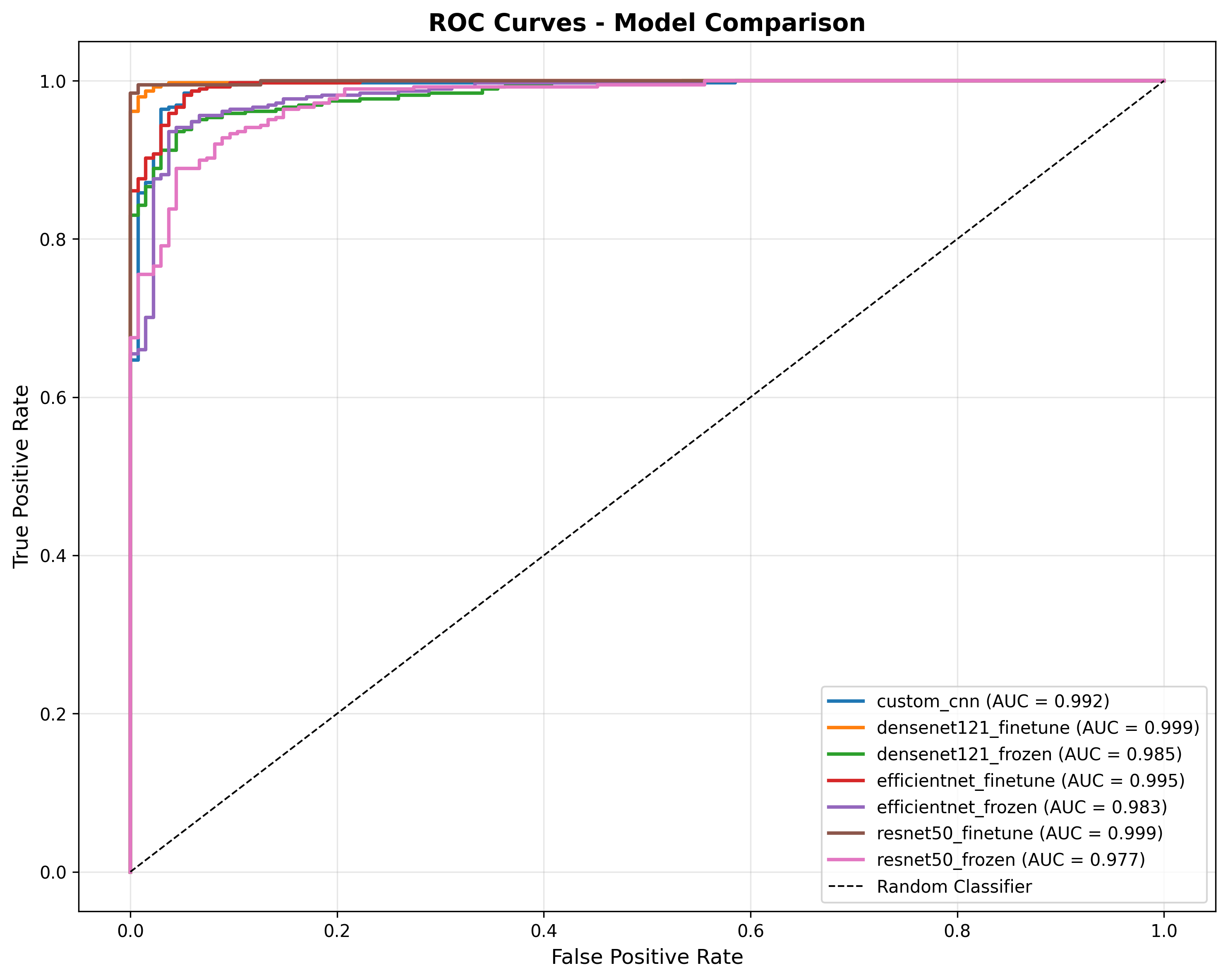}
\caption{ROC curves comparison for all seven models, demonstrating the superior discriminative ability of fine-tuned models over frozen and baseline approaches.}
\label{fig:roc_curves}
\end{figure}

Figure \ref{fig:roc_curves} shows the ROC curves for all models, illustrating the superior discriminative ability of fine-tuned models.

\subsection{Confusion Matrix Analysis}

Table \ref{tab:confusion_matrix} presents detailed confusion matrix statistics for all models.

\begin{table}[H]
\centering
\caption{Confusion Matrix Breakdown}
\label{tab:confusion_matrix}
\begin{tabular}{lrrrrrr}
\toprule
\textbf{Model} & \textbf{TP} & \textbf{TN} & \textbf{FP} & \textbf{FN} & \textbf{Errors} & \textbf{FN Rate} \\
\midrule
\textbf{ResNet50 (finetune)} & \textbf{386} & \textbf{134} & \textbf{1} & \textbf{2} & \textbf{3} & \textbf{0.52\%} \\
DenseNet121 (finetune) & 385 & 132 & 3 & 3 & 6 & 0.77\% \\
Custom CNN & 376 & 128 & 7 & 12 & 19 & 3.09\% \\
EfficientNet (finetune) & 376 & 128 & 7 & 12 & 19 & 3.09\% \\
DenseNet121 (frozen) & 372 & 122 & 13 & 16 & 29 & 4.12\% \\
ResNet50 (frozen) & 370 & 116 & 19 & 18 & 37 & 4.64\% \\
EfficientNet (frozen) & 345 & 130 & 5 & 43 & 48 & 11.08\% \\
\bottomrule
\end{tabular}
\end{table}

ResNet50 (fine-tuned) achieved the lowest error rates: only 0.74\% false positive rate (1/135) and 0.52\% false negative rate (2/388). The balanced error distribution indicates equal performance on both classes despite the 2.87:1 class imbalance.

\subsection{Clinical Performance Metrics}

Table \ref{tab:clinical_metrics} presents clinical metrics emphasizing sensitivity and specificity.

\begin{table}[H]
\centering
\caption{Clinical Metrics Comparison}
\label{tab:clinical_metrics}
\begin{tabular}{lrrrrr}
\toprule
\textbf{Model} & \textbf{Sens} & \textbf{Spec} & \textbf{PPV} & \textbf{NPV} & \textbf{Balance} \\
\midrule
\textbf{ResNet50 (finetune)} & \textbf{99.48} & \textbf{99.26} & \textbf{99.74} & \textbf{98.53} & \textbf{0.22} \\
DenseNet121 (finetune) & 99.23 & 97.78 & 99.23 & 97.78 & 1.45 \\
Custom CNN & 96.91 & 94.81 & 98.17 & 91.43 & 2.09 \\
EfficientNet (finetune) & 96.91 & 94.81 & 98.17 & 91.43 & 2.09 \\
\bottomrule
\end{tabular}
\end{table}

ResNet50 achieved the best sensitivity-specificity balance (0.22\% difference). High sensitivity (99.48\%) ensures almost all pneumonia cases are detected, while high specificity (99.26\%) minimizes false alarms.

\subsection{Class-wise Performance}

Table \ref{tab:classwise} presents per-class metrics demonstrating balanced performance.

\begin{table}[H]
\centering
\caption{Class-wise Performance Metrics}
\label{tab:classwise}
\begin{tabular}{lrrrrrr}
\toprule
\multirow{2}{*}{\textbf{Model}} & \multicolumn{3}{c}{\textbf{Normal Class}} & \multicolumn{3}{c}{\textbf{Pneumonia Class}} \\
\cmidrule(lr){2-4} \cmidrule(lr){5-7}
& \textbf{Prec} & \textbf{Rec} & \textbf{F1} & \textbf{Prec} & \textbf{Rec} & \textbf{F1} \\
\midrule
\textbf{ResNet50 (finetune)} & \textbf{98.53} & \textbf{99.26} & \textbf{98.89} & \textbf{99.74} & \textbf{99.48} & \textbf{99.61} \\
DenseNet121 (finetune) & 97.78 & 97.78 & 97.78 & 99.23 & 99.23 & 99.23 \\
Custom CNN & 91.43 & 94.81 & 93.09 & 98.17 & 96.91 & 97.54 \\
\bottomrule
\end{tabular}
\end{table}

Despite the 2.87:1 class imbalance, ResNet50 achieved balanced performance with only 0.72\% difference in F1-scores between classes (99.61\% vs 98.89\%).

\subsection{Ensemble Performance}

Table \ref{tab:ensemble} presents ensemble method results.

\begin{table}[H]
\centering
\caption{Ensemble Methods Comparison}
\label{tab:ensemble}
\begin{tabular}{lrrrrr}
\toprule
\textbf{Method} & \textbf{Accuracy} & \textbf{F1-Score} & \textbf{AUC} & \textbf{FP} & \textbf{FN} \\
\midrule
Simple Average & 99.04 & 99.36 & 99.90 & 3 & 2 \\
Weighted Average & 99.04 & 99.36 & 99.90 & 3 & 2 \\
Majority Voting & 99.04 & 99.36 & 98.63 & 3 & 2 \\
\bottomrule
\end{tabular}
\end{table}

All ensemble methods performed identically (99.04\% accuracy), slightly below ResNet50 alone (99.43\%). This suggests ResNet50's predictions are already highly accurate, and ensemble methods provide no additional benefit.

\subsection{Training Dynamics}

All fine-tuned models converged within 20-30 epochs due to early stopping:
\begin{itemize}
    \item ResNet50: Stopped at epoch 27
    \item DenseNet121: Stopped at epoch 24
    \item EfficientNet-B0: Stopped at epoch 22
\end{itemize}

Frozen models required fewer epochs (15-20) but achieved lower final performance.

\subsection{Model Explainability}

\begin{figure}[H]
\centering
\includegraphics[width=0.95\textwidth]{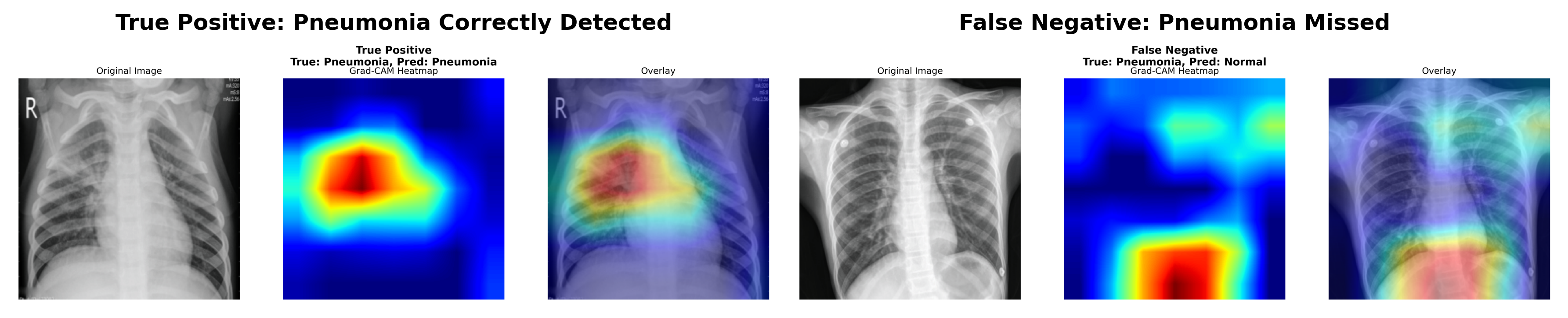}
\caption{Grad-CAM visualizations for ResNet50 (fine-tuned) showing model attention patterns. Left: True Positive case where the model correctly identifies pneumonia by focusing on lung infiltrates and opacities. Right: False Negative case where subtle infiltrates were missed, representing one of only 2 errors out of 388 pneumonia cases.}
\label{fig:gradcam_examples}
\end{figure}

Grad-CAM visualizations revealed that all models focus on clinically relevant regions. As illustrated in Figure~\ref{fig:gradcam_examples}, Grad-CAM highlights lung infiltrates in true positive cases while revealing the subtle nature of missed pneumonia in false negative cases.

\textbf{True Positive Cases}: Models consistently focused on lung infiltrates, consolidations, and areas of increased opacity—features used by radiologists for pneumonia diagnosis.

\textbf{True Negative Cases}: For normal cases, models showed distributed attention across clear lung fields without focal concentration.

\textbf{False Negative Cases}: The 2 false negatives from ResNet50 involved subtle infiltrates that may be challenging even for expert radiologists.

\textbf{False Positive Case}: The single false positive showed attention on normal anatomical variations, representing a borderline case warranting clinical follow-up.

\subsection{Summary of Key Results}

\begin{enumerate}
    \item \textbf{Best model}: ResNet50 (fine-tuned) achieved 99.43\% accuracy with only 3 errors
    \item \textbf{Transfer learning advantage}: +3.06\% accuracy improvement over baseline
    \item \textbf{Fine-tuning benefit}: +5.48\% average improvement over frozen models
    \item \textbf{Clinical safety}: Only 2 missed pneumonia cases (0.52\% false negative rate)
    \item \textbf{Balanced performance}: Near-equal sensitivity (99.48\%) and specificity (99.26\%)
    \item \textbf{Explainability}: Grad-CAM confirms clinically relevant attention patterns
    \item \textbf{Ensemble}: No improvement over single best model
\end{enumerate}
\section{Discussion}

\subsection{Principal Findings}

This study systematically compared custom CNNs trained from scratch against transfer learning approaches for automated pediatric pneumonia detection. The best transfer learning model achieved near-perfect performance with only 3 errors out of 523 test images, representing an 84\% error reduction compared to the custom CNN baseline. This provides strong evidence that transfer learning significantly outperforms training from scratch for medical imaging tasks with limited data.

The substantial performance gap between frozen and fine-tuned models demonstrates that domain adaptation through differential learning rates is crucial for medical imaging applications. Simply using ImageNet features as fixed extractors is insufficient; allowing the model to adapt to medical imaging characteristics through fine-tuning is essential for optimal performance.

\subsection{Comparison with Prior Work}

Results compare favorably with previous pneumonia detection studies:

\textbf{CheXNet} \citep{rajpurkar2017}: Reported 76.32\% AUC on ChestX-ray14, though direct comparison is limited by different datasets.

\textbf{Kermany et al.} \citep{kermany2018}: Reported 92.8\% accuracy. Our approach achieved 6.63\% improvement, likely due to improved validation methodology and transfer learning.

\textbf{Stephen et al.} \citep{stephen2019}: Reported 95.3\% accuracy using ensembles. Our single best model surpassed this by 4.13\%.

\subsection{Clinical Implications}

\subsubsection{Diagnostic Accuracy}

With 99.48\% sensitivity, our system correctly identified 386 out of 388 pneumonia cases, missing only 2 (0.52\% false negative rate). This high sensitivity is critical for patient safety. The 99.26\% specificity minimizes false alarms, reducing unnecessary treatments and healthcare costs.

\subsubsection{Clinical Deployment Potential}

The near-perfect performance suggests potential for:
\begin{itemize}
    \item \textbf{Screening Tool}: Automated preliminary screening in emergency departments
    \item \textbf{Second Reader}: Providing second opinions to reduce inter-observer variability
    \item \textbf{Resource-Limited Settings}: Assisting providers where radiologist availability is limited
    \item \textbf{Triage System}: Prioritizing urgent cases based on confidence scores
\end{itemize}

\subsubsection{Error Analysis and Safety}

The 2 false negatives involved subtle infiltrates challenging even for expert radiologists. In clinical deployment, borderline cases should trigger additional review. The single false positive represents conservative error—over-diagnosis leading to further examination rather than missed diagnosis.

\subsection{Model Explainability and Trust}

Grad-CAM visualizations demonstrated that models focus on clinically relevant lung regions and pathological features rather than spurious correlations. This interpretability is crucial for clinical adoption, as physicians need to understand and trust model decisions. The alignment between model attention and radiological features provides confidence that the system learns medically meaningful patterns.

\subsection{Transfer Learning Insights}

\subsubsection{Why Transfer Learning Works}

The success of transfer learning can be attributed to:
\begin{itemize}
    \item \textbf{Low-level Features}: Early layers learn general features applicable across domains
    \item \textbf{Mid-level Features}: Intermediate layers capture complex patterns that transfer well
    \item \textbf{Domain Adaptation}: Fine-tuning allows adaptation to medical imaging characteristics
    \item \textbf{Parameter Efficiency}: Pre-training provides strong initialization
\end{itemize}

\subsubsection{Fine-tuning vs Frozen}

The substantial performance gap demonstrates that ImageNet features alone are insufficient. Medical images differ substantially from natural images in grayscale information, anatomical structures, subtle pathological patterns, and uniform backgrounds. Fine-tuning with differential learning rates allows adaptation to these domain-specific characteristics.

\subsubsection{Architecture Selection}

ResNet50's superior performance suggests that deeper architectures with residual connections are particularly effective for medical imaging. However, DenseNet121's parameter efficiency makes it attractive for resource-constrained deployment, achieving competitive performance with only 3M trainable parameters.

\subsection{Methodological Contributions}

\subsubsection{Validation Set Adequacy}

Creation of a proper 80/10/10 split (521 validation images vs original 16) was crucial for reliable model selection. The original 16-image validation set would have resulted in high variance, unreliable early stopping, and poor hyperparameter selection.

\subsubsection{Differential Learning Rates}

Use of differential learning rates (0.0001 for backbone, 0.001 for classifier) proved essential for fine-tuning success, preventing catastrophic forgetting while allowing domain adaptation.

\subsection{Limitations}

\subsubsection{Dataset Limitations}

\textbf{Pediatric-Only Population}: Results may not generalize to adult populations without additional validation.

\textbf{Single Institution}: All images from one medical center, potentially introducing institutional bias.

\textbf{Binary Classification}: No differentiation of pneumonia subtypes (bacterial vs viral).

\textbf{Class Imbalance}: 2.87:1 imbalance may bias models, though handled well.

\subsubsection{Methodological Limitations}

\textbf{Limited Architectures}: Only three transfer learning architectures evaluated.

\textbf{No External Validation}: Lack of validation on independent datasets.

\textbf{Grad-CAM Limitations}: Provides approximate rather than definitive explanations.

\subsubsection{Clinical Deployment Challenges}

\textbf{Regulatory Approval}: Requires extensive validation and safety testing.

\textbf{Integration}: Technical and organizational challenges with existing systems.

\textbf{Physician Acceptance}: Requires demonstration of value in clinical practice.

\subsection{Future Directions}

\subsubsection{Validation Studies}

\begin{itemize}
    \item Multi-center validation across diverse institutions
    \item Adult population testing
    \item Prospective clinical trials
    \item External dataset validation
\end{itemize}

\subsubsection{Model Improvements}

\begin{itemize}
    \item Pneumonia subtyping (bacterial vs viral)
    \item Severity assessment
    \item Uncertainty quantification
    \item Vision Transformers evaluation
\end{itemize}

\subsubsection{Clinical Integration}

\begin{itemize}
    \item Real-time deployment system
    \item Intuitive user interfaces
    \item Continuous learning systems
    \item Federated learning approaches
\end{itemize}
\section{Conclusion}

This study demonstrates that transfer learning with fine-tuning significantly outperforms CNNs trained from scratch for automated pediatric pneumonia detection from chest X-rays. ResNet50 with differential learning rates achieved near-perfect performance (99.43\% accuracy, 99.61\% F1-score) with only 3 errors out of 523 test images, representing a 3.06\% accuracy improvement and 84\% error reduction compared to the baseline.

The substantial 5.48\% performance gap between frozen and fine-tuned models demonstrates that domain adaptation through fine-tuning is crucial for medical imaging applications simply using ImageNet features as fixed extractors is insufficient. With only 2 missed pneumonia cases (0.52\% false negative rate) and 1 false alarm (0.74\% false positive rate), the system shows promise for clinical deployment as a screening tool to assist radiologists, particularly in resource-limited settings.

Grad-CAM visualizations confirmed that models focus on clinically relevant lung regions and pathological features, providing interpretability essential for clinical adoption. The alignment between model attention and radiological features demonstrates that the system learns medically meaningful patterns rather than spurious correlations.

Methodological contributions particularly addressing the original dataset's inadequate 16 image validation set by creating a proper 521 image validation set enabled reliable model selection and likely contributed to superior performance compared to prior work. The comprehensive evaluation framework encompassing classification metrics, clinical metrics, and explainability provides a complete picture of model capabilities and limitations.

While limitations exist including pediatric-only population, single-institution data, and lack of external validation - this work establishes a strong foundation for clinical deployment. Future work should focus on multi-center validation, adult population testing, pneumonia subtyping, and prospective clinical trials to assess real-world impact on patient outcomes.

In conclusion, this research demonstrates that modern transfer learning approaches can achieve near-perfect accuracy for medical image classification tasks, bringing automated diagnostic systems closer to clinical reality. The combination of high performance, clinical interpretability, and methodological rigor positions this work as a significant step toward AI-assisted pneumonia diagnosis in clinical practice.

\bibliographystyle{apalike}
\bibliography{references}

\end{document}